\documentclass[runningheads]{llncs}
\usepackage[T1]{fontenc}
\usepackage{graphicx}
\usepackage{caption}
\usepackage{subcaption}

\usepackage{multirow}

\usepackage{enumitem}

\usepackage{nameref}

\usepackage{xcolor}
\usepackage{soul}

\begin{document}
\title{UniCausal: Unified Benchmark and Repository \\for Causal Text Mining}
\titlerunning{UniCausal}
%

\author{Fiona Anting Tan\inst{1}\orcidID{0000-0002-2828-1831} \and
Xinyu Zuo\inst{2} \and
See-Kiong Ng\inst{1}}
\authorrunning{Tan et al.}

\institute{Institute of Data Science, National University of Singapore, Singapore \email{tan.f@u.nus.edu, seekiong@nus.edu.sg} \and
Tencent Technology, China
\email{xylonzuo@tencent.com}
\\\url{https://github.com/tanfiona/UniCausal}
}

\maketitle

\begin{abstract}
Current causal text mining datasets vary in objectives, data coverage, and annotation schemes. These inconsistent efforts prevent modeling capabilities and fair comparisons of model performance. Furthermore, few datasets include cause-effect span annotations, which are needed for end-to-end causal relation extraction. To address these issues, we propose UniCausal, a unified benchmark for causal text mining across three tasks: (I) Causal Sequence Classification, (II) Cause-Effect Span Detection and (III) Causal Pair Classification. We consolidated and aligned annotations of six high quality, mainly human-annotated, corpora, resulting in a total of 58,720, 12,144 and 69,165 examples for each task respectively. Since the definition of causality can be subjective, our framework was designed to allow researchers to work on some or all datasets and tasks. To create an initial benchmark, we fine-tuned BERT pre-trained language models to each task, achieving 70.10\% Binary F1, 52.42\% Macro F1, and 84.68\% Binary F1 scores respectively.


\end{abstract}
\section{Introduction}
\begin{figure*}[!h]
  \centering    
  \includegraphics[width=0.8\textwidth]{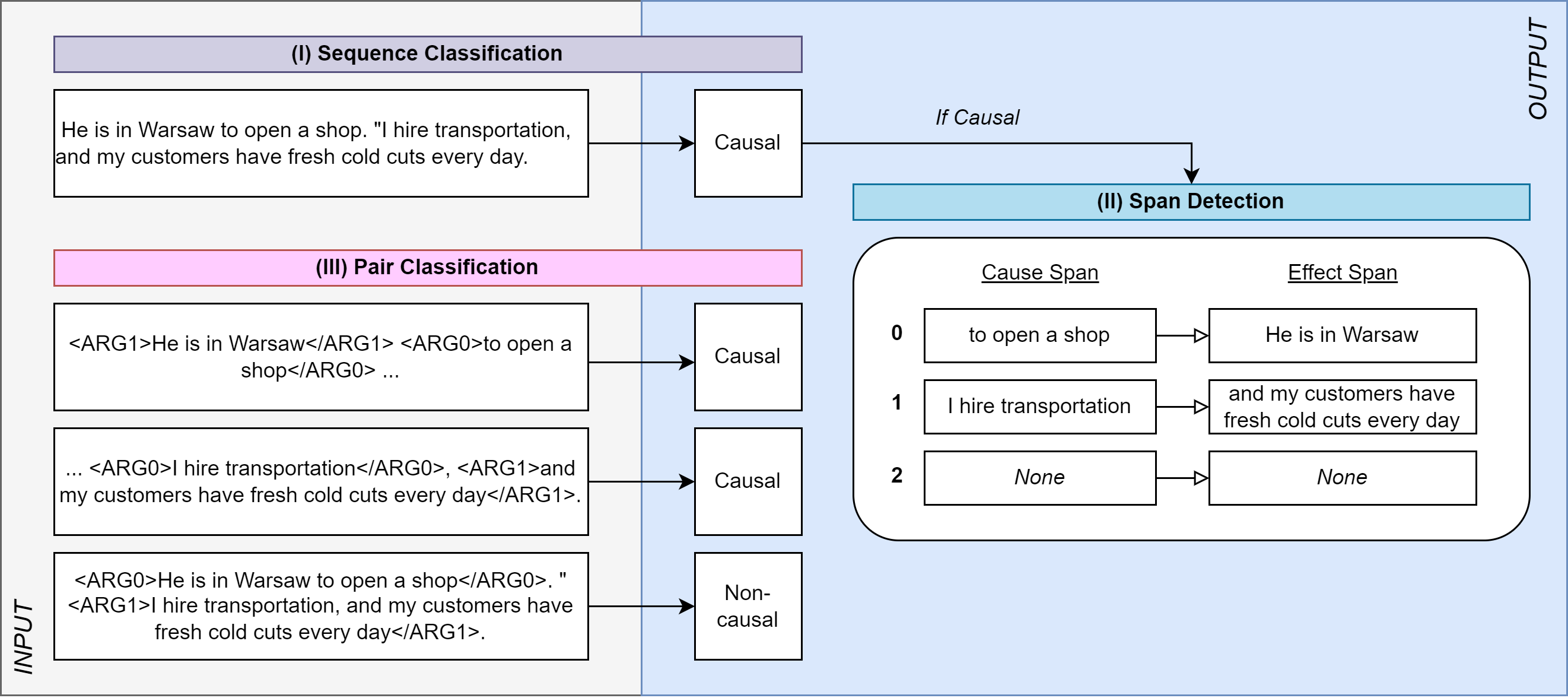}
  \caption{A two-sentence example that contains causal relations. (I) Sequence Classification aims to return a \emph{Causal} label. (II) Span Detection identifies the text related to the Cause-Effect spans. This example contains three annotated relation pairs, where two are labeled \emph{Causal} and one is labeled \emph{Non-causal}. These are target labels of the (III) Pair Classification task.}
  \label{fig:tasks}
\end{figure*}

Causal text mining relates to the extraction of causal information from text. Given an input text, we are interested to know if and where causal information occurs. Researchers can use extracted causal information as a knowledge base \cite{DBLP:conf/cikm/HeindorfSWNP20,DBLP:conf/kr/LuoSZHW16,DBLP:conf/ijcai/LiD0HD20}, for summarization, or prediction \cite{DBLP:conf/wsdm/RadinskyH13}. Since causality is an important part of human cognition, causal text mining has important natural language understanding applications \cite{girju-2003-automatic,stasaski-etal-2021-automatically,dunietz-etal-2020-test}. Figure \ref{fig:tasks} illustrates three causal text mining tasks (Sequence Classification, Span Detection and Pair Classification) and their expected output. 

Currently, large and diverse causal text mining corpora are limited \cite{DBLP:journals/corr/Asghar16a}. Across datasets, annotation guidelines vary \cite{yang2022survey}. These issues hinder both modeling capabilities and fair comparisons between models. Additionally, working on independent tasks and datasets runs the risk of training task-specialized and dataset-specific models that are not generalizable. 


The contributions of UniCausal are as follows:
\begin{itemize}
    \item To the best of our knowledge, we are the first to produce a unified benchmark and resource for causal text mining. Apart from consolidating six datasets for three tasks, we also employed competitive natural language processing (NLP) models to obtain baseline scores, reported in this paper.
    \item Our framework provides a seamless way for researchers to design individual or joint models, while benchmarking their performance against clearly defined test sets across some or all the processed corpora.
    \item Our codes and processed data is available online. Our trained baseline model checkpoints are uploaded to Huggingface Hub.~\footnote{Our repository is at \url{https://github.com/tanfiona/UniCausal}. Links to our trained baseline models are available in the repository.}
\end{itemize}

The rest of the paper is organized as follows: Section \ref{sec:related} performs a literature review. Section \ref{sec:benchmark} describes how we created the consolidated corpus while Section \ref{sec:modeling} outlines our baseline models. Subsequently, Section \ref{sec:results} discusses our findings and Section \ref{sec:conclusion} concludes.

\section{Related Work}
\label{sec:related}

In this paper, we are interested in supporting end-to-end causal relation extraction. More specifically, given an input sequence, a model should be designed to detect whether the sequence contains causal relations, and if so, where its causal arguments are.

\subsection{Tasks}

In the earlier days, researchers extracted causal relations and assessed the validity of the extracted relations directly \cite{girju-2003-automatic,DBLP:journals/dke/IttooB13a}. In the recent years, a two-stepped approach is increasingly popular: After the successful identification of causal sequences (Causal Classification), detection of the cause-effect spans can be conducted on the positive sequences (Cause-Effect Span Detection). This step-wise approach is practiced by shared tasks \cite{mariko-etal-2020-financial,mariko-etal-2021-financial,tan-etal-2022-causal,tan-etal-2022-event}.

\subsection{Datasets}
Research in causal text mining has been limited by data deficiency issues \cite{yang2022survey}. The lack of standardized datasets hinders comparisons in performance across models \cite{DBLP:journals/corr/Asghar16a}. In the next few paragraphs, we describe some common causal text mining datasets.

AltLex \cite{hidey-mckeown-2016-identifying}\footnote{\url{https://github.com/chridey/altlex}} investigates causal relations with alternative lexicalizations (AltLex) connectives in single sentences. AltLex connectives are an open-class of markers with varied linguistic constructions \cite{webber2019penn}, like \emph{``so (close) that''} and \emph{``This may help explain why''}. The limitations of the AltLex dataset is that it (1) is small in size, (2) ignores explicit and implicit signals in causal relations, (3) ignores inter-sentence causal relations, and (4) assumes Cause and Effect spans to be all the words before and after the signals.

BECAUSE 2.0 \cite{dunietz-etal-2017-corpus}\footnote{\url{https://github.com/duncanka/BECAUSE/tree/2.0}} contains annotations for causal language in single sentences. Cause, Effect or Connective spans were annotated based on principles of Construction Grammar. Documents and articles were selected from four data sources: Congressional Hearings from 2014 NLP Unshared Task in PoliInformatics (CHRG), Penn Treebank (PTB), Manually Annotated Sub-Corpus (MASC), and the New York Times Annotated Corpus (NYT). The limitations of BECAUSE 2.0 is that (1) it is relatively small in size and (2) ignores inter-sentence causal relations.

CausalTimeBank (CTB)~\cite{mirza-etal-2014-annotating,mirza-tonelli-2014-analysis}\footnote{\url{https://github.com/paramitamirza/Causal-TimeBank}} annotated only explicit causal relations in the TempEval-3 based on rule-based algorithm. EventStoryLine (ESL)~\cite{caselli-vossen-2017-event}\footnote{\url{https://github.com/tommasoc80/EventStoryLine}} annotated both explicit and implicit causal relations in the Event Coreference Bank+. Both CTB and ESL includes annotations of intra- and inter-sentence causal relations between events. These two datasets are popular amongst researchers studying Event Causality Identification (ECI)~\cite{zuo-etal-2021-improving,zuo-etal-2020-knowdis,cao-etal-2021-knowledge}, which aims to classify if a pair of events are causal or not given its context. However, these two datasets have limited usage outside the event text mining space because they only annotate events, and furthermore, exclude the context of the event in the argument span.

For Penn Discourse Treebank V3.0 (PDTB)~\cite{webber2019penn}\footnote{\url{https://catalog.ldc.upenn.edu/LDC2019T05}} and SemEval 2010 Task 8 (SemEval) \cite{hendrickx-etal-2010-semeval}, causal relations were not the original focus of the dataset. Causal relations are one out of the many relations they annotated. PDTB annotated discourse relations between arguments, expressed either explicitly, implicitly or in AltLex forms. The main limitation of PDTB is that causal relations expressed within clauses are not annotated. SemEval was annotated for the purpose of semantic relations classification. They accepted only noun phrases with common-noun heads as relation arguments. SemEval's limitations are: (1) the context is not included in the argument span and (2) it also ignores inter-sentence causal relations.

\begin{table*}[htbp]
\centering
\resizebox{\textwidth}{!}{
\begin{tabular}{lp{4cm}cll}
\hline
Corpus & Source & Inter-sent & Linguistic & Arguments \\\hline
AltLex   \cite{hidey-mckeown-2016-identifying} & News & No & AltLex & Words before/after signal \\
BECAUSE 2.0   \cite{dunietz-etal-2017-corpus} & News, Congress Hearings & No & Explicit & Phrases \\
CausalTimeBank (CTB)   \cite{mirza-etal-2014-annotating} & News & Yes & All & Event head word(s) \\
EventStoryLine V1.0 (ESL)   \cite{caselli-vossen-2017-event} & News & Yes & All & Event head word(s) \\
Penn Discourse Treebank V3.0   (PDTB) \cite{webber2019penn} & News & Yes & All & Clauses \\
SemEval 2010 Task 8 (SemEval)   \cite{hendrickx-etal-2010-semeval} & Web & No & All & Noun phrases\\\hline
\end{tabular}
}
\caption{Six popular causal corpora and their annotation coverage in terms of: data source, sentence lengths, linguistic construction and argument types.}\label{tab:differences}
\end{table*}

Table \ref{tab:differences} summarizes the differences across the six datasets. The current disarray across causal text mining datasets leads to three key missed opportunities: Firstly, for training, researchers are unable to seamlessly increase models' exposure to a wide range of examples. Secondly, for evaluation, researchers cannot make fair comparisons of model performance with one another. Thirdly, it is inconvenient for researchers to test their models' generalizability to other corpora. To address these issues, we propose UniCausal, a large consolidated resource of annotated texts for causal text mining. We relied on the above six high quality corpora and aligned each corpus' definitions, where possible, to cater to the three causal text mining tasks. With the exception of CTB, all other datasets were annotated by humans.




\subsection{Other large causal resources}

Although some large corpora or knowledge bases (KBs) that include causal relations already exists, they are annotated in a semi-supervised manner and constructed using rule-based methods.~\footnote{E.g.: Bootstrapped versions of AltLex \cite{hidey-mckeown-2016-identifying} and SCITE \cite{DBLP:journals/ijon/LiLZR21}; Causal KBs: CauseNet \cite{DBLP:conf/cikm/HeindorfSWNP20}, CausalNet \cite{DBLP:conf/kr/LuoSZHW16} and CausalBank \cite{DBLP:conf/ijcai/LiD0HD20}; Semantic KBs that include causal relations: FrameNet \cite{ruppenhofer2016framenet} and ConceptNet \cite{DBLP:conf/aaai/SpeerCH17}}. For example, CauseNet \cite{DBLP:conf/cikm/HeindorfSWNP20} detected causal relations automatically using causal dependency path patterns obtained in a boostrap fashion. Examples from such corpora are of lower quality and have less linguistic variation. Thus, although they are useful databases for common causal relations, they contribute minimally to training and fair testing of models' reasoning abilities. We perform a short study on this matter in Section \ref{ssec:causenet}. Different from them, our corpus aims to capture linguistic, syntactic and semantic variation. Furthermore, by training a text mining model on our extensive corpus, researchers can potentially create an even larger causal KB by extracting more relations compared to rule-based methods.


\section{Methodology}


\subsection{Creation of UniCausal}
\label{sec:benchmark}

\subsubsection{Causal Text Mining Datasets}
\label{sec:datasets}

We combine six datasets that can be used for causal text mining: AltLex \cite{hidey-mckeown-2016-identifying}, BECAUSE 2.0 \cite{dunietz-etal-2017-corpus}, CTB \cite{mirza-etal-2014-annotating,mirza-tonelli-2014-analysis}, ESL \cite{caselli-vossen-2017-event}, PDTB \cite{webber2019penn} and SemEval \cite{hendrickx-etal-2010-semeval}. 

\subsubsection{Causal Text Mining Tasks}
\label{sec:tasks}

There are three causal text mining tasks that we focus on, corresponding to the tasks shown in Figure \ref{fig:tasks}:
\begin{enumerate}[label=(\Roman*)]
    \item Sequence Classification: Given an example, does it contain any causal relations?
    \item Span Detection: Given a causal example, which words in the input text correspond to the Cause and Effect arguments? Identify up to three causal relations and their spans.
    \item Pair Classification: Given the marked argument or entity pair, are they causally related such that the first argument (\texttt{ARG0}) causes the second argument (\texttt{ARG1})?
\end{enumerate}
Since pairs can be \emph{Non-causal}, we marked the arguments with \texttt{ARG0} and \texttt{ARG1} instead of \texttt{CAUSE} and \texttt{EFFECT}. 

\subsubsection{Data Processing}
\label{ssec:preprocessing}

For every dataset, we only focus on examples that were of three or shorter sentences. For \emph{Causal} sequences, only the sentences that contain the arguments were retained. We split the dataset into train and test sets based on previous works' recommendations, or if not, randomly. Finally, we process each dataset to fit into the format required for our three tasks described below.

Let a unique sequence of text be represented by a vector $\vec{w} = w_1, w_2, ..., w_N$ of $N$ word tokens. Each sequence has a binary label $s$ of either 1 or 0, representing \emph{Causal} or \emph{Non-causal} respectively.

\begin{enumerate}[label=(\Roman*)]

    \item Sequence Classification: Worked on both \emph{Causal} and \emph{Non-causal} texts. Each example text is unique with a target label, $s$.
    
    \item Span Detection: Worked only on \emph{Causal} texts. Each example text is unique, and at the current stage, we focus on examples with up to three cause-effect relations only. We approach the task as a token classification task, where the annotated spans in the texts were converted to BIO-format (Begin (\texttt{B}), Inside (\texttt{I}), Outside (\texttt{O})) \cite{ramshaw-marcus-1995-text} for two types (Cause (\texttt{C}), Effect (\texttt{E})). Therefore, there were five possible labels per word: \texttt{B-C}, \texttt{I-C}, \texttt{B-E}, \texttt{I-E} and \texttt{O}. The corresponding target token vector is $\vec{t} = t_1, t_2, ..., t_N$, where each $t_n$ represents one of the five labels. For examples with multiple relations, we sort them based on the location of the \texttt{B-C}, followed by \texttt{B-E} if tied. See Figure \ref{fig:tasks}'s spans for example. For multiple causal relations, $\vec{w}$ has multiple token vectors $\vec{t^v}$ where $v=0,1,2$, since we permit up to three causal relations per unique text.
    
    \item Pair Classification: Worked on both \emph{Causal} and \emph{Non-causal} texts. Each example text is unique after taking into account of where special tokens \texttt{ARG0} and \texttt{ARG1} are located. For a sequence of text with $N$ word tokens, we include $2 \cdot a$ special beginning and end tokens\footnote{(\texttt{<ARG0>}, \texttt{</ARG0>}) marks the boundaries of a Cause span, while (\texttt{<ARG1>}, \texttt{</ARG1>}) marks the boundaries of a corresponding Effect span.} such that the input word vector $\vec{u}$ is now of length $N+2 \cdot a$. $a$ represents the number of arguments in the example. $\vec{w}$ can have multiple versions of $\vec{u}$, depending on the location of the special tokens. For example, in Figure \ref{fig:tasks}, there are three Pair Classification examples for one Sequence Classification example.
\end{enumerate}

\begin{table}[]
\centering
\resizebox{0.7\columnwidth}{!}{
\begin{tabular}{l|p{10mm}|cc|c|cc}
\hline
 & & \multicolumn{2}{c|}{(I) Seq} & \multicolumn{1}{c|}{(II) Span} & \multicolumn{2}{c}{(III) Pair} \\\cline{3-7} 
Corpus & Split & \emph{Non-causal} & \emph{Causal} & \emph{Causal} & \emph{Non-causal} & \emph{Causal} \\\hline
AltLex & Train & 277 & 300 & 300 & 296 & 315 \\
 & Test & 286 & 115 & 115 & 289 & 127 \\\hline
BEC- & Train & 183 & 716 & 716 & 266 & 902 \\
AUSE & Test & 10 & 41 & 41 & 14 & 46 \\\hline
CTB & Train & 1,651 & 234 & \multicolumn{1}{c|}{-} & 3,047 & 270 \\
 & Test & 274 & 42 & \multicolumn{1}{c|}{-} & 444 & 48 \\\hline
ESL & Train & 957 & 1,043 & \multicolumn{1}{c|}{-} & \multicolumn{1}{c}{-} & \multicolumn{1}{c}{-} \\
 & Test & 119 & 113 & \multicolumn{1}{c|}{-} & \multicolumn{1}{c}{-} & \multicolumn{1}{c}{-} \\\hline
PDTB & Train & 24,901 & 8,917 & 8,917 & 32,587 & 9,809 \\
 & Test & 5,796 & 2,055 & 2,055 & 7,694 & 2,294 \\\hline
Sem- & Train & 6,976 & 999 & \multicolumn{1}{c|}{-} & 6,997 & 1,003 \\
Eval & Test & 2,387 & 328 & \multicolumn{1}{c|}{-} & 2,389 & 328 \\\hline
\multicolumn{2}{c|}{Total} & 43,817 & 14,903 & 12,144 & 54,023 & 15,142\\\hline
\end{tabular}
}
\caption{Size of dataset. ``-" indicates tasks not applicable to the corpus.}\label{tab:final_data}
\end{table}

The final data sizes are reflected in Table \ref{tab:final_data}. The number of Span Detection example tallies with the positive instances of Sequence Classification examples because multiple cause-effect relation spans (i.e. $\vec{t^0}, \vec{t^1}$ and $\vec{t^2}$) were grouped into a unique example (i.e. same $\vec{w}$), which we term as a `grouped' example. At evaluation, performance metrics were calculated at the `ungrouped' level so that every causal relation is evaluated against equally.

Since each data source has a different data format, our codes had to extract the text sequences and relations from different annotation types: Our codes work for BECAUSE's `brat', ESL's `CAT' and CTB's `TimeML'/`XML' data formats. PDTB use their own standoff annotations format. AltLex and SemEval datasets are more user-friendly, in that they are stored in `CSV' and `JSON' formats, and can be interpreted directly in a single file. Due to the brevity of space, we describe how we handle the different annotation guidelines and our data processing steps for each corpus in detail in our Supplementary Material online. We also upload the data processing codes for each source to our repository. The final, post-processed datasets are all stored in `CSV' for convenience. We also built a custom dataset loader based on Huggingface's \texttt{load\_dataset} function, such that users only need to indicate the datasets of interest  either as a list of inputs within the script (E.g. `\texttt{dataset\_name=[`altlex',`because']}'), or directly in the command line (E.g. `\texttt{$--$dataset\_name altlex because}').



\begin{table*}[]
\centering
\resizebox{0.98\textwidth}{!}{
\begin{tabular}{lp{10mm}p{10mm}p{10mm}p{10mm}p{10mm}p{10mm}p{10mm}p{10mm}p{10mm}p{10mm}p{10mm}}\hline
 & \multicolumn{4}{c}{(I) Sequence Classification} & \multicolumn{3}{c}{(II) Span Detection} & \multicolumn{4}{c}{(III) Pair Classification} \\\cline{2-12}
\multicolumn{1}{c}{Test Set} & \multicolumn{1}{c}{P} & \multicolumn{1}{c}{R} & \multicolumn{1}{c}{F1} & \multicolumn{1}{c}{Acc} & \multicolumn{1}{c}{P} & \multicolumn{1}{c}{R} & \multicolumn{1}{c}{F1} & \multicolumn{1}{c}{P} & \multicolumn{1}{c}{R} & \multicolumn{1}{c}{F1} & \multicolumn{1}{c}{Acc} \\\hline

All  & 71.13 \textpm0.80 & {69.14 \textpm1.60} & {70.10 \textpm0.58} & {86.27 \textpm0.15} & 46.33 \textpm1.22 & 60.35 \textpm0.30 & 52.42 \textpm0.90 & 85.44 \textpm0.96 & {83.93 \textpm0.44} & {84.68 \textpm0.27} & {93.68 \textpm0.16} \\
 
AltLex  & 50.76 \textpm1.61 & {63.48 \textpm4.60} & 56.37 \textpm2.49 & 71.87 \textpm1.19 & 27.74 \textpm1.20 & {42.99 \textpm0.85} & 33.72 \textpm1.12 & 82.60 \textpm1.99 & {87.09 \textpm1.53} & {84.76 \textpm0.66} & {90.43 \textpm0.55} \\
 
BECAUSE  & {92.32 \textpm1.69} & {70.24 \textpm2.04} & {79.77 \textpm1.68} & {71.37 \textpm2.24} & 32.51 \textpm2.82 & 44.30 \textpm2.33 & 37.47 \textpm2.57 & 87.93 \textpm1.73 & {94.78 \textpm1.94} & {91.21 \textpm1.18} & {86.00 \textpm1.90} \\
 
CTB  & 42.37 \textpm2.11 & {66.19 \textpm4.26} & 51.58 \textpm1.82 & 83.48 \textpm1.21 & \multicolumn{1}{c}{-} & \multicolumn{1}{c}{-} & \multicolumn{1}{c}{-} & 75.66 \textpm3.61 & {72.50 \textpm6.81} & {73.94 \textpm4.68} & {95.04 \textpm0.78} \\
 
ESL  & 76.11 \textpm2.04 & {67.43 \textpm3.45} & {71.45 \textpm1.89} & 73.79 \textpm1.34 & \multicolumn{1}{c}{-} & \multicolumn{1}{c}{-} & \multicolumn{1}{c}{-} & \multicolumn{1}{c}{-} & \multicolumn{1}{c}{-} & \multicolumn{1}{c}{-} & \multicolumn{1}{c}{-} \\
 
PDTB  & 72.59 \textpm0.61 & {66.34 \textpm1.63} & {69.31 \textpm0.70} & 84.63 \textpm0.17 & 47.77 \textpm1.22 & 61.54 \textpm0.29 & 53.78 \textpm0.88 & 84.56 \textpm1.17 & {82.04 \textpm0.46} & {83.28 \textpm0.36} & {92.43 \textpm0.23} \\
 
SemEval  & {73.39 \textpm1.18} & {89.51 \textpm1.59} & {80.64 \textpm0.46} & {94.81 \textpm0.16} & \multicolumn{1}{c}{-} & \multicolumn{1}{c}{-} & \multicolumn{1}{c}{-} & 93.38 \textpm0.88 & {96.10 \textpm0.59} & {94.71 \textpm0.23} & {98.70 \textpm0.07} \\\hline
 
\end{tabular}
}
\caption{Mean and
standard deviation of performance metrics for different test sets across the three tasks, across five random seeds. All models were trained on all six datasets, where applicable. Tasks that are not applicable to the dataset are indicated by ``-". Scores are reported in percentages (\%).seeds.}\label{tab:extended_results2}
\end{table*}
\begin{table*}[]
\centering
\resizebox{\textwidth}{!}{
\begin{tabular}{p{10mm}lp{45mm}p{45mm}p{10mm}p{10mm}p{10mm}p{10mm}}
\multicolumn{8}{l}{(I) Sequence Classification}\\\hline
{Corpus} & {Source} & {Features} & {Model} & {P} & {R} & {F1} & {Acc} \\\hline
AltLex & \cite{hidey-mckeown-2016-identifying} & Lexical & Support Vector Machine & \textbf{61.98} & 58.51 & \textbf{60.19} & 67.68 \\
 & Ours (All) & BERT & BERT+LR & 50.76 & \textbf{63.48} & 56.37 & \textbf{71.87} \\
 & Ours (AltLex) & BERT & BERT+LR & 50.58 & 53.57 & 51.85 & 71.52 \\
\hline
BEC- & \cite{zuo-etal-2020-towards}$\mathsection$ $\ddagger$ & Discourse, SA & PSAN & - & - & 81.70 & - \\
AUSE & Ours (All) & BERT & BERT+LR & \textbf{92.32} & 70.24 & 79.77 & 71.37 \\
 & Ours (BECAUSE) & BERT & BERT+LR & 86.20 & \textbf{96.01} & \textbf{90.77} & \textbf{84.31} \\
\hline
CTB & \cite{kyriakakis-etal-2019-transfer}\^{} & n-gram & LR & 100.00 & 22.22 & 36.36 & - \\
 &  & word2vec & BIGRUATT & 67.04 & 73.89 & 69.98 & - \\
 &  & ELMO & BIGRUATT & \textbf{81.29} & 70.28 & 75.08 & - \\
 &  & BERT & BERT+LR & 71.17 & \textbf{93.33} & \textbf{80.55} & - \\
 &  & BERT & BERT+BIGRUATT & 74.52 & 86.94 & 80.06 & - \\
 & Ours (All) & BERT & BERT+LR & 42.37 & 66.19 & 51.58 & 83.48 \\
 & Ours (CTB) & BERT & BERT+LR & 71.46 & 58.57 & 63.65 & \textbf{91.27} \\
\hline
ESL & \cite{kyriakakis-etal-2019-transfer}\^{} & n-gram & LR & \textbf{100.00} & 27.27 & 42.86 & - \\
 &  & word2vec & BIGRUATT & 70.09 & 60.91 & 63.65 & - \\
 &  & ELMO & BIGRUATT & 77.47 & 59.09 & 66.55 & - \\
 &  & BERT & BERT+LR & 62.44 & 87.17 & 72.35 & - \\
 &  & BERT & BERT+BIGRUATT & 66.15 & 83.64 & 73.09 & - \\
 & Ours (All) & BERT & BERT+LR & 76.11 & 67.43 & 71.45 & 73.79 \\
 & Ours (ESL) & BERT & BERT+LR & 75.90 & \textbf{87.79} & \textbf{81.21} & \textbf{80.17} \\
\hline
PDTB & \cite{ponti-korhonen-2017-event}$\ddagger$ & Lexical & Shallow CNN & 39.80 & 75.29 & 52.04 & 63.00 \\
 &  & Lexical & FFNN & 42.04 & 71.74 & 53.01 & 66.44 \\
 &  & Lexical, positional, event & FFNN & 42.37 & \textbf{76.45} & 54.52 & 66.35 \\
 & \cite{zuo-etal-2020-towards}$\mathsection$ $\ddagger$ & Discourse, SA & PSAN & - & - & \textbf{76.60} & - \\
 & \cite{tan-EtAl:2022:LREC}$\mathsection$ & BERT & BERT+LR & - & - & 74.45 & - \\
 & Ours (All) & BERT & BERT+LR & 72.59 & 66.34 & 69.31 & 84.63 \\
 & Ours (PDTB) & BERT & BERT+LR & \textbf{73.54} & 67.35 & 70.31 & \textbf{85.11} \\
\hline
Sem- & \cite{DBLP:conf/icdm/NikiSIM19} & n-gram & Random Forest & - & - & 52.80 & - \\
Eval &  & n-gram & LR & - & - & 81.90 & - \\
 &  & word2vec & LSTM & - & - & 85.60 & - \\
 &  & word2vec & LSTM + Self-Attention & - & - & 86.90 & - \\
 & \cite{kyriakakis-etal-2019-transfer}\^{} & n-gram & LR & 88.67 & 66.83 & 76.22 & - \\
 &  & word2vec & BIGRUATT & 93.96 & 87.59 & 90.64 & - \\
 &  & ELMO & BIGRUATT & \textbf{94.45} & 91.26 & \textbf{92.81} & - \\
 &  & BERT & BERT+LR & 86.62 & 97.09 & 91.55 & - \\
 &  & BERT & BERT+BIGRUATT & 86.80 & \textbf{96.63} & 91.45 & - \\
 & Ours (All) & BERT & BERT+LR & 73.39 & 89.51 & 80.64 & 94.81 \\
 & Ours (SemEval) & BERT & BERT+LR & 87.84 & 91.40 & 89.58 & \textbf{97.43}\\\hline
 \\
\multicolumn{8}{l}{(III) Pair Classification}\\\hline
{Corpus} & {Source} & {Features} & {Model} & {P} & {R} & {F1} & {Acc} \\\hline
Sem- & \cite{9028985} & Lexical, semantic, dependency & Bayesian Classifier & - & - & 66.00 & 93.00 \\
Eval &  & word2vec & CNN & - & - & 66.00 & 88.00 \\
 &  & GrammarTags & CNN & - & - & 86.60 & 93.00 \\
 & Ours (All) & BERT & BERT+LR & 93.38 & \textbf{96.10} & 94.71 & 98.70 \\
 & Ours (SemEval) & BERT & BERT+LR & \textbf{93.96} & 95.67 & \textbf{94.80} & \textbf{98.73}\\\hline
\end{tabular}
}
\caption{Evaluation metrics for each dataset in the literature review compared to our benchmark (Ours). We do not cover methods that rely on the connectives as features for Classification tasks. Notations: \^{} Rebalanced the dataset, $\mathsection$ Evaluated on k-folds or different folds, $\ddagger$ Slightly different definitions for class labels. Abbreviations: Self-Attention Embeddings (SA), Logistic Regression (LR), Bidirectional GRU + Self-Attention (BIGRUATT), Feed-forward neural network (FFNN)
}\label{tab:litreview}
\end{table*}

\subsection{Baseline Model}
\label{sec:modeling}

Transformer-based pre-trained language models are the state-of-the-art in NLP. To create our initial benchmark, we used pre-trained Bidirectional Encoder Representations from Transformers (BERT) models \cite{devlin-etal-2019-bert}. First, sequences are tokenized into token embeddings ($\vec{r_n}$). Special start (\texttt{[CLS]}) and end (\texttt{[SEP]}) tokens were added to the input sequence. For the Pair Classification task only, we added four special tokens to the vocabulary to represent the boundaries of the two arguments. The BERT encoder is fine-tuned to our task during training.

\subsubsection{Sequence and Pair Classification}\label{ssec:model_seq}
For each classification task, we pooled the token embeddings into a sequence embedding by extracting the embedding on the \texttt{[CLS]} token. The sequence embeddings was then fed into a sequence classifier $g(.)$ to predict logits for the two labels: \emph{Causal} and \emph{Non-causal}, as in $\hat{y} = g(\vec{r_{[CLS]}})$. We compared the logits with the true sequence label $y$ to calculate Cross-Entropy (CE) Loss for learning.

\begin{equation}
\mathcal{L}_{seq}= -y \cdot \log(\hat{y}) - (1-y) \cdot \log(1-\hat{y})
\end{equation}

\subsubsection{Span Detection}\label{ssec:model_span}
We fed the token embeddings into a token classifier ($f(.)$), as in $\hat{t_{n}} = f(\vec{r_n})$. The classifier returns predicted logits for the five BIO-CE token labels, which aims to predict the cause-effect span in the input sequence, obtained via \texttt{argmax}. Note that given the current simple set up, the span detection model can only predict one cause-effect relation per input sequence. Again, the logits and true token vector $\vec{t^v}$ were used to calculate CE Loss.

\begin{equation}
\mathcal{L}_{token}= -\sum_{n=1}^{N} \sum_{i=1}^{C}
t_{n,i} \cdot \log(\hat{t_{n,i}})
\end{equation}

\subsubsection{Evaluation Metrics}
For the two Classification tasks, we calculated the Accuracy (Acc), Precision (P), Recall (R) and F1 scores per experiment. For Span Detection, we referred to evaluation metrics from earlier Cause-Effect Span Detection \cite{mariko-etal-2020-financial,mariko-etal-2021-financial,tan-etal-2022-event} shared tasks, and used the Macro P, R and F1 metrics. The token classification evaluation scheme by \texttt{seqeval} \cite{seqeval,ramshaw-marcus-1995-text} reverts the BIO-formatted labels to the original form (i.e. Cause (\texttt{C}) and Effect (\texttt{E})) for evaluation. Our default evaluation scripts report metrics for all and each corpus. In the next section, we present the average and standard deviation scores obtained from multiple runs using five random seeds. 

\section{Experiments}
\label{sec:results}
In this section, we describe experiments performed on the UniCausal corpus.

\subsection{Baseline Performance}
\label{ssec:results_performance}

In Table \ref{tab:extended_results2}, we present the performance of the baseline BERT models when trained on all datasets, and tested on all and each dataset. Across all test sets, baseline models achieved 70.10\% Binary F1 score for Sequence Classification, 52.42\% Macro F1 score for Span Detection, and 84.68\% Binary F1 score for Pair Classification. 

Overall, regardless of the dataset, performance for Pair Classification is always better than Sequence Classification. F1 scores for Span Detection is poor in comparison to the Classification tasks. This finding correlates with the difficulty of each task: For Pair Classification, since the prompts that already identifies the arguments are provided, it is arguably a simpler task than Sequence Classification. For Span Detection, it is much more challenging than both Classification tasks because it involves accurate identification of the words that corresponds to the cause and effect, not just the mere identification that they exist. Furthermore, the baseline token classification set-up was too simplistic, and unable to handle multiple cause-effect span relations in the same sequence. For each input text, only one pair of Cause and Effect will be predicted. Thus, if multiple relation exists, only one pair can be predicted correctly at best.

In Table \ref{tab:litreview}, we provide a snapshot of evaluation metrics reported by previous works on the datasets. It is challenging to make claims about model superiority from this table alone, since different papers used different train-test splits and some papers altered the dataset composition by rebalancing it. Nevertheless, for datasets like AltLex and SemEval, the development set was predefined by the dataset creators. Thus, comparisons between previous work and ours can be made concretely. For Sequence Classification with AltLex, \cite{hidey-mckeown-2016-identifying}'s handcrafted lexical features fed through a Support Vector Machine achieved an F1 score of 60.19\%, beating us. For Sequence Classification with SemEval, our best F1 score of 89.58\% surpasses methods covered by \cite{DBLP:conf/icdm/NikiSIM19} which, at best, achieved 86.90\% using word2vec embeddings fed through a Long-Short Term Memory Self-Attention network. Finally, for Pair Classification with SemEval, our BERT-based model consistently surpasses Bayesian Classifier and Convolutional Neural Network methods explored by \cite{9028985}.

All in all, our baseline model is simple but competitive. From Table \ref{tab:litreview} alone, it is apparent that the Causal Text Mining community lacks a consistent way to benchmark performance. Therefore, we hope that from here on, the scores presented in Table \ref{tab:extended_results2} will serve as an initial, universal baseline score for the Causal Text Mining community to beat.

\subsection{Impact of Datasets}
\label{ssec:results_datasets}

In Table \ref{tab:matrix}, we present the F1 scores when training and testing on different corpus. This table reflects how compatible each corpus is to one another. When testing on all the datasets, we noticed that training on all datasets returned the best performance across all tasks by a large margin. Training on any one dataset was unable to achieve similar performance. Meanwhile, the generalized model trained on all datasets did not always return the best performance for each corpus. Given the differences in definitions and linguistic coverage of each dataset, it is expected that for some datasets, specializing on its own data distributions leads to better performance. However, such specialized models are more likely to overfit and lack generalizability. Thus, good performance on one dataset but not others should be handled with caution.

\begin{table*}[]
\centering

\resizebox{1\textwidth}{!}{
\begin{tabular}{p{21mm}p{25mm}p{25mm}p{25mm}p{25mm}p{25mm}p{25mm}p{25mm}}
\multicolumn{8}{l}{(I) Sequence Classification}\\\hline
\multicolumn{1}{l}{} & \multicolumn{7}{c}{Test Set} \\ \cline{2-8} 
\multicolumn{1}{l}{Training Set} & \multicolumn{1}{l}{All} & \multicolumn{1}{l}{AltLex} & \multicolumn{1}{l}{BECAUSE} & \multicolumn{1}{l}{CTB} & \multicolumn{1}{l}{ESL} & \multicolumn{1}{l}{PDTB} & \multicolumn{1}{l}{SemEval} \\ \hline
All & \textbf{70.10 \textpm0.58} & \textbf{56.37 \textpm2.49} & 79.77 \textpm1.68 & 51.58 \textpm1.82 & 71.45 \textpm1.89 & 69.31 \textpm0.70 & 80.64 \textpm0.46 \\
AltLex & 32.93 \textpm3.57*** & 51.85 \textpm2.53 & 36.47 \textpm11.18*** & 38.21 \textpm6.20* & 53.30 \textpm8.37** & 22.91 \textpm5.79*** & 55.83 \textpm6.68*** \\
BECAUSE & 39.15 \textpm0.99*** & 47.02 \textpm1.52** & \textbf{90.77 \textpm2.22***} & 25.17 \textpm1.34*** & 63.49 \textpm1.94** & 42.49 \textpm0.68*** & 23.71 \textpm1.93*** \\
CTB & 33.49 \textpm5.48*** & 55.91 \textpm7.63 & 54.73 \textpm9.40** & \textbf{63.65 \textpm5.55**} & 33.26 \textpm15.44** & 25.97 \textpm3.73*** & 51.76 \textpm13.85** \\
ESL & 39.62 \textpm0.89*** & 46.29 \textpm1.15** & 90.12 \textpm1.05*** & 30.84 \textpm1.35*** & \textbf{81.21 \textpm2.35***} & 42.55 \textpm1.25*** & 26.15 \textpm2.62*** \\
PDTB & 60.99 \textpm0.76*** & 48.94 \textpm1.88** & 69.61 \textpm2.16** & 39.54 \textpm1.88*** & 38.71 \textpm3.15*** & \textbf{70.31 \textpm0.56*} & 19.75 \textpm3.35*** \\
SemEval & 28.25 \textpm0.86*** & 28.95 \textpm1.74*** & 16.91 \textpm3.40*** & 38.51 \textpm3.44** & 45.95 \textpm3.50*** & 10.11 \textpm1.61*** & \textbf{89.58 \textpm0.71***}\\\hline

\\
\multicolumn{5}{l}{(II) Span Detection}\\\cline{1-5}
\multicolumn{1}{l}{} & \multicolumn{4}{c}{Test Set} \\ \cline{2-5}
\multicolumn{1}{l}{Training Set} & \multicolumn{1}{l}{All} & \multicolumn{1}{l}{AltLex} & \multicolumn{1}{l}{BECAUSE} & \multicolumn{1}{l}{PDTB} & \multicolumn{1}{l}{} & \multicolumn{1}{l}{} & \multicolumn{1}{l}{}\\ \cline{1-5}
All & \textbf{52.42 \textpm0.90} & \textbf{33.72 \textpm1.12} & 37.47 \textpm2.57 & 53.78 \textpm0.88 &  &  &  \\
AltLex & 6.20 \textpm0.74*** & 21.45 \textpm1.87*** & 11.51 \textpm1.63*** & 5.47 \textpm0.76*** &  &  &  \\
BECAUSE & 12.74 \textpm0.35*** & 7.38 \textpm2.19*** & \textbf{37.79 \textpm5.77} & 12.60 \textpm0.34*** &  &  &  \\
PDTB & 51.97 \textpm0.48 & 6.73 \textpm0.94*** & 35.84 \textpm2.42 & \textbf{55.02 \textpm0.38*} &  &  & \\\cline{1-5}

\\
\multicolumn{7}{l}{(III) Pair Classification}\\\cline{1-7}
\multicolumn{1}{l}{} & \multicolumn{6}{c}{Test Set} \\ \cline{2-7} 
\multicolumn{1}{l}{Training Set} & \multicolumn{1}{l}{All} & \multicolumn{1}{l}{AltLex} & \multicolumn{1}{l}{BECAUSE} & \multicolumn{1}{l}{CTB} & \multicolumn{1}{l}{PDTB} & \multicolumn{1}{l}{SemEval} & \multicolumn{1}{l}{} \\\cline{1-7}
All & \textbf{84.68 \textpm0.27} & \textbf{84.76 \textpm0.66} & \textbf{91.21 \textpm1.18} & \textbf{73.94 \textpm4.68} & 83.28 \textpm0.36 & 94.71 \textpm0.23 &  \\
AltLex & 31.83 \textpm3.93*** & 80.57 \textpm2.48* & 48.44 \textpm20.00** & 20.06 \textpm7.14*** & 25.11 \textpm8.75*** & 57.72 \textpm14.52** &  \\
BECAUSE & 36.40 \textpm0.64*** & 47.99 \textpm1.33*** & 90.01 \textpm1.95 & 23.58 \textpm1.52*** & 38.39 \textpm0.37*** & 25.23 \textpm2.02*** &  \\
CTB & 20.17 \textpm5.78*** & 19.16 \textpm15.64*** & 22.00 \textpm10.92*** & 73.29 \textpm6.14 & 7.02 \textpm6.06*** & 63.69 \textpm5.65*** &  \\
PDTB & 68.13 \textpm0.88*** & 40.34 \textpm1.52*** & 82.59 \textpm2.17*** & 26.74 \textpm2.42*** & \textbf{83.70 \textpm0.34} & 33.64 \textpm1.76*** &  \\
SemEval & 26.66 \textpm1.86*** & 37.07 \textpm6.58*** & 25.70 \textpm11.46*** & 50.63 \textpm1.74*** & 8.08 \textpm3.20*** & \textbf{94.80 \textpm0.28} &\\\cline{1-7}

\end{tabular}
}
\caption{Mean and
standard deviation of F1 score across different training and test set combinations across five random seeds. Scores are reported in percentages (\%). For each panel, the top score per column is bolded. Paired T-test was conducted against the first row per panel, where all datasets were used for training. Statistical significance: ***$<0.001$, **$<0.01$, *$<0.05$.}\label{tab:matrix}
\end{table*}




\subsection{Adding CauseNet to investigate the importance of linguistic variation in examples}
\label{ssec:causenet}

Researchers might want to incorporate custom data to train and test their models. The modular structure of our code, in terms of data loading and evaluation, allows this. To illustrate, we obtained around 50,000 training and 5,000 testing causal examples from CauseNet \cite{DBLP:conf/cikm/HeindorfSWNP20} suitable for Pair Classification. After data processing, it is straightforward to load causenet with other datasets by calling it with the command `\texttt{$--$dataset\_name}'. 

In terms of F1, we only found improvements in performance for CTB (77.08\%) and AltLex (85.04\%), which are rule-based and relatively template-based respectively. Unsurprisingly, the model also performs perfectly on CauseNet. BECAUSE (89.80\%), PDTB (83.20\%), and SemEval (94.31\%) had poorer performance. Despite incorporating a large number of rule-based causal examples, the model does not learn about the semantics of causal relations. We believe this is due to the lack of linguistic variation covered. This again motivates our focus to include mainly human-annotated data in UniCausal for both better training and fairer testing.

\section{Conclusion}
 \label{sec:conclusion}
We propose UniCausal, a unified resource and benchmark for causal text mining. Our codes were designed to allow researchers to work on some or all datasets and tasks, while still comparing their performance fairly against us or others. Researchers can easily include new datasets too. In this paper, we provided evaluation metrics per dataset as an initial benchmark for future researchers to compete against. 

We hope to see researchers use UniCausal to design joint models that concurrently learn from multiple causal text mining tasks and datasets. A unified model will be able to collaboratively learn about causality from various objectives and knowledge sources, to be more universally adaptable and generalizable to unseen examples. 

For future work, we intend to include more datasets relevant to causal text mining, like the Son Facebook dataset \cite{son-etal-2018-causal}, FinCausal \cite{mariko-etal-2020-financial,mariko-etal-2021-financial}, and Causal News Corpus \cite{tan-etal-2022-causal,tan-etal-2022-event}. We would also expand UniCausal to include longer examples with much more sentences. Finally, we will replicate more models to include in our benchmark.


%
%
%
\bibliographystyle{splncs04}
\bibliography{custom, anthology}

\end{document}